\documentclass[11pt]{article}
\usepackage{hyperref}
 
\usepackage{amssymb,amsmath,color,times}

\usepackage[dvips]{graphicx}
 \author{
Francis Bach \\
\url{francis.bach@mines.org}\\
INRIA - WILLOW Project-Team \\
Laboratoire d'Informatique de l'Ecole Normale Sup\'erieure \\
(CNRS/ENS/INRIA UMR 8548) \\
45, rue d'Ulm,
75230 Paris, France
  }
\title{Exploring Large Feature Spaces \\ with Hierarchical Multiple Kernel Learning}
\def \anc{ {\rm A} }
\def \hull{ {\rm hull} }
\def \sou{ {\rm sources} }
\def \des{ {\rm D} }

\def \X { \mathcal{X}}
\def \F { \mathcal{F}}

\newcommand{\BEAS}{\begin{eqnarray*}}
\newcommand{\EEAS}{\end{eqnarray*}}
\newcommand{\BEA}{\begin{eqnarray}}
\newcommand{\EEA}{\end{eqnarray}}
\newcommand{\BEQ}{\begin{equation}}
\newcommand{\EEQ}{\end{equation}}
\newcommand{\BIT}{\begin{itemize}}
\newcommand{\EIT}{\end{itemize}}
\newcommand{\BNUM}{\begin{enumerate}}
\newcommand{\ENUM}{\end{enumerate}}
\newcommand{\BA}{\begin{array}}
\newcommand{\EA}{\end{array}}

\newcommand{\Diag}{\mathop{\rm Diag}}

\newcommand{\var}{\mathop{ \rm var}}

\newcommand{\rb}{\mathbb{R}}
\newcommand{\BlackBox}{\rule{1.5ex}{1.5ex}}  

\newenvironment{proof}{\par\noindent{\bf Proof\ }}{\hfill\BlackBox\\[2mm]}

\newtheorem{proposition}{Proposition}
\newtheorem{lemma}{Lemma}

\newcommand{\mysec}[1]{Section~\ref{sec:#1}}
\newcommand{\eq}[1]{Eq.~(\ref{eq:#1})}
\newcommand{\myfig}[1]{Figure~\ref{fig:#1}}

\begin{document}
\maketitle

\begin{abstract}
For supervised and unsupervised learning, positive definite kernels allow to use large and potentially infinite dimensional feature spaces with a computational cost that only depends on the number of observations. This is usually done through the penalization of predictor functions by  Euclidean or Hilbertian norms. In this paper, we explore penalizing by sparsity-inducing norms such as the $\ell^1$-norm or the block $\ell^1$-norm. We assume that the kernel decomposes into a large sum of individual basis kernels which can be embedded in a directed acyclic graph; we show that it is then possible to perform kernel selection through a hierarchical multiple kernel learning framework,  in polynomial time in the number of selected kernels. This framework is naturally applied to non linear variable selection; our extensive simulations on synthetic datasets and datasets from the UCI repository show that efficiently exploring the large feature space through sparsity-inducing norms leads to state-of-the-art predictive performance.
\end{abstract}

 \section{Introduction}

In the last two decades,  kernel methods have been a prolific  theoretical and algorithmic machine learning framework. By using appropriate regularization by Hilbertian norms, representer theorems enable to consider large and potentially infinite-dimensional feature spaces while working within an implicit feature space no larger than the number of observations. This has led to numerous works on kernel design adapted to specific data types and generic kernel-based algorithms for many learning tasks (see, e.g., \cite{smola-book,Cristianini2004}).

Regularization by sparsity-inducing norms, such as the $\ell^1$-norm has also attracted a lot of interest in recent years. While early work has focused on efficient algorithms to solve the convex optimization problems, recent research has looked at the model selection properties and predictive performance of such methods, in the linear case~\cite{Zhaoyu} or within the multiple kernel learning framework~\cite{grouplasso}.

In this paper, we aim to bridge the gap between these two lines of research by trying to use $\ell^1$-norms \emph{inside} the feature space. Indeed, feature spaces are large and we expect the estimated predictor function to require only a small number of features, which is exactly the situation where $\ell^1$-norms have proven advantageous. This leads to two natural questions that we try to answer in this paper: (1) Is it feasible to perform optimization in this very large feature space with cost which is polynomial in the size of the input space? (2) Does it lead to better predictive performance and feature selection?

 More precisely, we consider a positive definite kernel that can be expressed as a large sum of positive definite \emph{basis} or \emph{local kernels}. This exactly corresponds to the situation where a large feature space is the concatenation of  smaller feature spaces, and we aim to do selection among these many kernels, which may be done through  multiple kernel learning~\cite{skm}. One major difficulty however is that the number of these smaller kernels is usually exponential in the dimension of the input space and applying multiple kernel learning directly in this decomposition would be intractable.
 
In order to peform selection efficiently, we make the extra assumption that these small kernels can be embedded in a \emph{directed acyclic graph} (DAG). Following~\cite{cap, marie}, we consider in \mysec{mkl} a specific combination of $\ell^2$-norms that is adapted to the DAG, and will restrict the authorized sparsity patterns; in our specific kernel framework, we are able to use the DAG to design an optimization algorithm which has polynomial complexity in the number of selected kernels (\mysec{optimization}).  In simulations (\mysec{simulations}), we focus on  \emph{directed grids}, where our framework allows to perform non-linear variable selection.
We provide extensive experimental validation of our novel regularization framework; in particular, we compare it to the regular $\ell^2$-regularization and shows that it is always competitive and often leads to better performance, both on synthetic examples, and standard regression and classification datasets from the UCI repository.

Finally, we extend in \mysec{consistency} some of the known consistency results of the Lasso and multiple kernel learning~\cite{Zhaoyu,grouplasso}, and give a partial answer to the model selection capabilities of our regularization framework by giving necessary and sufficient conditions for model consistency. In particular, we show that our framework is adapted to estimating consistently only the \emph{hull}  of the relevant variables. Hence, by restricting the statistical power of our method, we gain computational efficiency.

\section{Hierarchical multiple kernel learning (HKL) }
\label{sec:mkl}

We consider the problem of predicting a random variable $Y \in \mathcal{Y} \subset \rb$ from a random variable $X \in \mathcal{X}$, where $\mathcal{X}$ and $\mathcal{Y}$ may be quite general spaces. We assume that we are given  $n$ i.i.d.~observations $(x_i,y_i) \in \mathcal{X} \times
\mathcal{Y}$, $ i =1,\dots,n$.
We define the \emph{empirical risk} of a function $f$ from $\mathcal{X}$ to 
$\rb$ as $\frac{1}{n} \sum_{i=1}^n \ell(y_i, f(x_i))$,
where $\ell: \mathcal{Y}\times \rb \mapsto \rb^+$ is a \emph{loss function}. We only assume that $\ell$ is     convex with respect to the second parameter (but not necessarily differentiable). Typical examples of loss functions  
are the square loss for regression, i.e., $\ell(y,\hat{y}) = \frac{1}{2}(y-\hat{y})^2$ for $y \in \rb$, and the logistic loss $\ell(y,\hat{y}) = \log(1+e^{-y\hat{y}})$ or the hinge loss
$\ell(y,\hat{y}) = \max\{0,1-y\hat{y}\}$ for binary classification, where $y \in \{-1,1\}$, leading respectively to logistic regression and support vector machines.
Other losses may be used for other settings (see, e.g.,~\cite{Cristianini2004} or the Appendix).

\subsection{Graph-structured positive definite kernels}

We assume that we are given a \emph{positive definite kernel} $k:\X \times \X \to \rb$, and that this kernel can be expressed as the sum, over an index set $V$, of basis kernels $k_v$, 
$v\in V$, i.e, for all $x,x' \in \X$,
$k(x,x') = \sum_{v \in V} k_v(x,x')$. For each $v\in V$, we   denote by $\mathcal{F}_v$ and $\Phi_v$ the 
feature space and feature map of $k_v$, i.e., for all $x,x' \in \X$,
 $k_v(x,x') = \langle \Phi_v(x), \Phi_v(x') \rangle$. Throughout the paper, we   denote by  $\| u \|$ the Hilbertian norm of $u$ and by
 $\langle u,v \rangle$ the associated dot product, where the precise space is omitted and can always be   inferred from the context.

Our sum assumption corresponds to a situation where the feature map $\Phi(x)$ and feature space $\mathcal{F}$  for $k$ is the \emph{concatenation} of the feature maps $\Phi_v(x)$ for each kernel $k_v$, i.e, 
$\F = \prod_{v \in V} \F_v$ and $\Phi(x) = ( \Phi_v(x) )_{v \in V}$. Thus, looking for a certain $\beta \in \F$ and a predictor function
$f(x) = \langle \beta, \Phi(x) \rangle$ is equivalent to looking jointly for
$\beta_v \in \F_v$, for all $v \in V$, and $f(x) = \sum_{v \in V} 
 \langle \beta_v , \Phi_v(x) \rangle$.

\begin{figure}
\begin{center}

\vspace*{-.2cm}

\includegraphics[scale=.43]{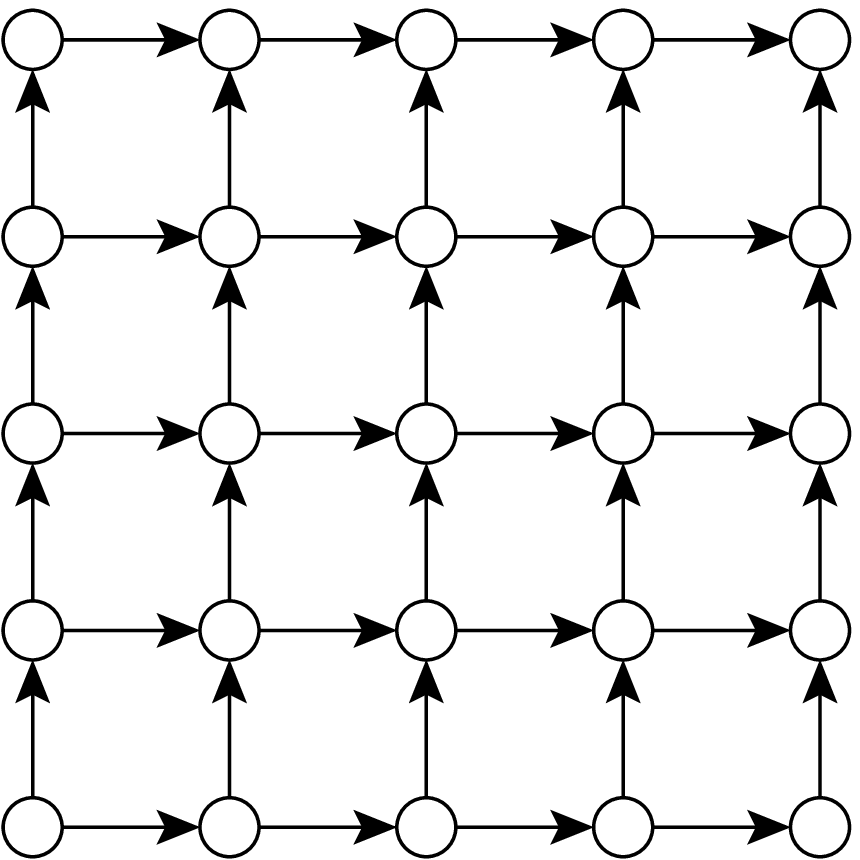} \hspace*{1.5cm}
\includegraphics[scale=.43]{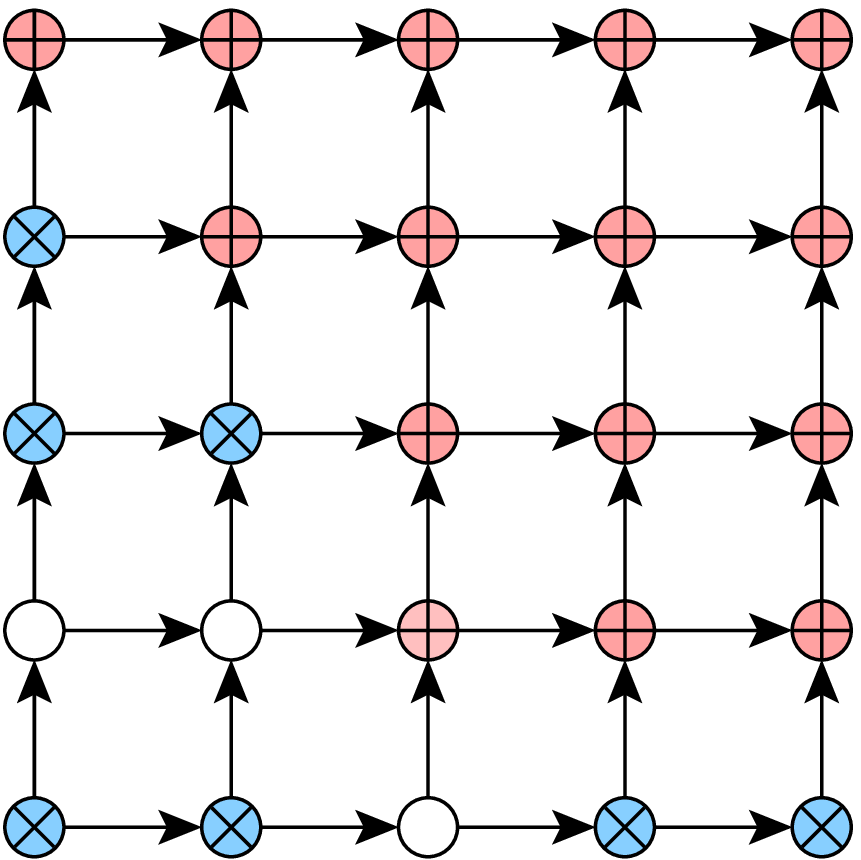} \hspace*{1.5cm}
\includegraphics[scale=.43]{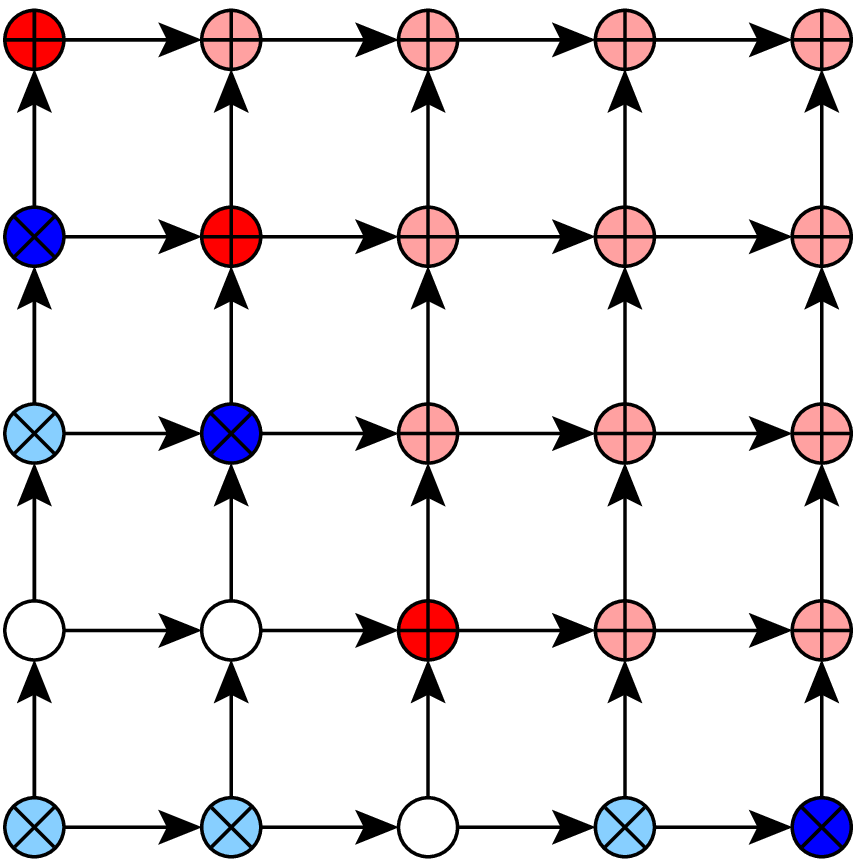}  \end{center}

\vspace*{-.5cm}

\caption{Example of graph and associated notions. (Left) Example of a 2D-grid. (Middle) Example of sparsity pattern ($\times$ in light blue) and the complement of its hull ($+$ in light red). (Right) Dark blue points ($\times$) are extreme points of the set of all active points (blue $\times$); dark red points ($+$) are the sources of the set of all red points ($+$).
}
\label{fig:2d}
\end{figure}

 As mentioned earlier, we make the assumption that the set $V$ can be embedded into a \emph{directed acyclic graph}. Directed acyclic graphs (referred to as DAGs) allow to naturally define the 
notions of \emph{parents}, \emph{children}, \emph{descendants} and \emph{ancestors}.  
Given a node $w \in V$, we   denote by $\anc(w) \subset V$ the set of its ancestors,
and by $\des(w)\subset V$, the set of its descendants. We use the convention that any $w$ is a descendant and an ancestor of itself, i.e., $w \in \anc(w)$ and $w \in \des(w)$.
 Moreover, for $W \subset V$, we let denote $\sou(W)$ the set of \emph{sources}  of the graph $G$ restricted to $W$ (i.e., nodes in  $W$ with no parents belonging to $W$).  Given a subset of nodes $W \subset V$, we can define the \emph{hull} of $W$ as the union of all ancestors of $w \in W$, i.e.,
$\hull(W) = \bigcup_{w \in W} A(w)$. Given a set $W$, we define the set of \emph{extreme points} of $W$ as the smallest subset $T \subset W $ such that $\hull(T) = \hull(W)$ (note that it is always well defined, as $\bigcap_{T \subset V, \
\hull(T) = \hull(W)} T$). See \myfig{2d} for   examples of these notions.

The goal of this paper is to perform kernel selection among the kernels $k_v$, $v \in V$. We essentially use the graph to limit the search to specific subsets of $V$. 
Namely, instead of considering all possible subsets of active (relevant) vertices, we are only interested in estimating correctly the hull of these relevant vertices;   in \mysec{graphreg}, we design a specific sparsity-inducing norms adapted to hulls.

In this paper, we primarily focus on kernels that can be expressed as ``products of sums'', and on the associated $p$-dimensional directed grids, while noting that our framework is applicable to many other kernels. Namely, we assume that the input space $\X$ factorizes into $p$ components $\X = \X_1 \times \cdots  \times \X_p$ and that
we are given $p$ sequences of length $q+1$ of kernels
$k_{ij}(x_i,x_i')$, $i \in \{1,\dots,p\}$, $j \in \{0,\dots,q\}$, such that
$k(x,x') = \sum_{j_1,\dots,j_p=0}^q \prod_{i=1}^p k_{ij_i}(x_i,x_i')
= \prod_{i=1}^p  \left(\sum_{j_i=0}^q k_{ij_i}(x_i,x_i') \right)
$. We thus have   a sum of $(q+1)^p$ kernels, that can be computed efficiently as a product of $p$ sums.
A natural DAG on $V = \prod_{i=1}^p \{0,\dots,q\}$
is defined by connecting each $(j_1,\dots,j_p)$
to $(j_1\!+\!1,j_2,\dots,j_p)$, $\dots$,
$(j_1,\dots,j_{p-1}, j_p\! +\!1)$. As shown in \mysec{graphreg}, this DAG will correspond to the constraint of selecting a given product of kernels only after all the subproducts are selected.
Those DAGs are especially suited to nonlinear variable selection, in particular with the polynomial and Gaussian kernels. In this context, products of kernels correspond to interactions between certain variables, and our DAG implies that we select an interaction only after all sub-interactions were already selected.

\textbf{Polynomial kernels} \hspace*{.2cm} We consider $\X_i = \rb$, $k_{ij}(x_i,x_i') ={ q \choose j}  (x_i x_i')^{j}$;  the full kernel is then equal to $k(x,x') = \prod_{i=1}^p \sum_{j=0}^q { q \choose j}  (x_i x_i')^j
= \prod_{i=1}^p ( 1 + x_i x_i')^q$. Note that this is not exactly the usual polynomial kernel (whose
feature space is the space of multivariate polynomials of \emph{total} degree less than $q$), since our kernel considers  polynomials  of \emph{maximal} degree $q$.

\textbf{Gaussian kernels} \hspace*{.2cm} We also consider $\X_i = \rb$, and the Gaussian-RBF kernel $  e^{-b(x -x')^2}$. The following decomposition is   
 the eigendecomposition of the non centered covariance operator for a normal distribution with variance $1/4a$ (see, e.g.,~\cite{williams00effect}):
$$ \textstyle 
e^{-b(x-x')^2}
=   \sum_{k=0}^{\infty} \frac{(b/A)^k}{2^k k!} [ e^{-\frac{b}{A}(a+c) x^2  }
H_k( \sqrt{2c} x) ]
[ e^{-\frac{b}{A}(a+c) (x')^2  }
H_k( \sqrt{2c} x') ],
$$ 
where $c^2 = a^2 + 2ab$, $A = a+b+c$, and $H_k$ is the $k$-th Hermite polynomial. By appropriately
truncating the sum, i.e, by considering that the first $q$ basis kernels are obtained from the first $q$ single Hermite polynomials, and the $(q+1)$-th kernel is summing over all other kernels, we obtain a decomposition of a uni-dimensional Gaussian kernel into $q+1$ components ($q$ of them are one-dimensional, the last one is infinite-dimensional, but can  be computed by differencing).
The decomposition ends up being close to a polynomial kernel of infinite degree, modulated by an exponential~\cite{Cristianini2004}. One may also use an \emph{adaptive} decomposition using kernel PCA (see, e.g.,~\cite{Cristianini2004,smola-book}), which is equivalent to using the eigenvectors of the empirical covariance operator associated with the data (and not the population one associated with the Gaussian distribution with same variance). In simulations, we tried both with no significant differences.

Finally, by taking product over all variables, we obtain a decomposition of the $p$-dimensional Gaussian kernel into $(q+1)^p$ components, that are adapted to nonlinear variable selection. Note that for $q=1$, we obtain ANOVA-like decompositions~\cite{Cristianini2004}.

\textbf{Kernels or features?} \hspace*{.2cm}
In this paper, we emphasize the \emph{kernel view}, i.e., we are given a kernel (and thus a feature space) and we explore it using $\ell^1$-norms. Alternatively, we could use the \emph{feature view}, i.e., we have a large structured set of features that we try to select from; however, the techniques developed in this paper assume that (a) each feature might be infinite-dimensional and (b) that we can sum all the local kernels efficiently (see in particular \mysec{reduced}). Following the kernel view thus seems slightly more natural.

\subsection{Graph-based structured regularization}

\label{sec:graphreg}
Given $\beta \in \prod_{v \in V} \mathcal{F}_v$, the natural Hilbertian norm
$\|\beta\|$ is defined through
$\|\beta\|^2 = \sum_{v \in V } \| \beta_v \|^2$. Penalizing with this norm is efficient because
summing all kernels $k_v$ is assumed feasible in polynomial time and we can bring to bear the usual kernel machinery; however, it does not lead to sparse solutions, where many $\beta_v$ will be exactly equal to zero.

As said earlier, we are only interested in the hull of the selected elements $\beta_v\in \F_v$,
$v \in V$; the hull of a set $I$
is characterized by the set of $v$, such that $\des(v) \subset I^c$, i.e., such that all descendants of $v$ are in the complement $I^c$: $\hull(I)= \{ v \in V, \des(v) \subset I^c\}^c$. 
Thus, if we try to estimate $\hull(I)$, we need to determine which $v \in V$ are such that
$\des(v) \subset I^c$. In our context, we   are hence looking at selecting  vertices $v \in V$ for which $\beta_{\des(v)} = (\beta_w)_{w \in \des(v)} = 0$.
 
We thus consider the following  structured block $\ell^1$-norm defined as
$$\sum_{v \in V} d_v \| \beta_{\des(v)} \|
= \sum_{v \in V} d_v  ( \sum_{w \in \des(v)} \| \beta_w\|^2  )^{1/2}
,$$
where $(d_v)_{v \in V}$ are positive weights.
Penalizing by such a norm will indeed impose that some of the vectors
$\beta_{\des(v)} \in \prod_{w \in \des(v)} \mathcal{F}_w$ are exactly zero. We thus consider the following minimization problem\footnote{Following~\cite{skm}, we consider the square of the norm, which does not change the regularization properties, but allow simple links with multiple kernel learning.}:
 \BEQ
\label{eq:milasso2}
\textstyle
\min_{ \beta \in \prod_{v\in V}\! \mathcal{F}_v}
\frac{1}{n} \sum_{i=1}^n \ell(y_i,  \sum_{v \in V} \langle \beta_v , \Phi_v(x_i) \rangle )
+ \frac{\lambda }{2} \left(  \sum_{v \in V}  d_v \| \beta_{\des(v)}  \| \right)^2.
\EEQ
Our Hilbertian norm is a Hilbert space instantiation of the hierarchical norms recently introduced by~\cite{cap}. If all Hilbert spaces are finite dimensional, our particular choice of norms corresponds to an  ``$\ell^1$-norm of $\ell^2$-norms''. While with uni-dimensional groups or kernels, the ``$\ell^1$-norm of $\ell^\infty$-norms'' allows an efficient path algorithm for the square loss and when the DAG is a tree~\cite{cap}, this is not possible anymore with groups of size larger than one, or when the DAG is a not a tree. In \mysec{optimization}, we propose a novel algorithm  to solve the  associated optimization problem in time polynomial in the number of selected groups or kernels, for all group sizes, DAGs and losses. Moreover, in \mysec{consistency}, we show under which conditions a solution to the problem in \eq{milasso2} consistently estimates the hull of the sparsity pattern.

Finally, note that in certain settings (finite dimensional Hilbert spaces and distributions with absolutely continuous densities), these norms have the effect of selecting a given kernel  \emph{only after  all of its ancestors}~\cite{cap}. This is another explanation why hulls end up being selected, since to include a given vertex in the models, the entire set of ancestors must also be selected.

\section{Optimization problem}
\label{sec:optimization}

In this section, we give optimality conditions  for the
problems in  \eq{milasso2}, as well as optimization algorithms with polynomial time complexity in the number of selected kernels. In simulations we consider total numbers of kernels larger than $10^{30}$, and thus such efficient algorithms are essential to the success of hierarchical multiple kernel learning (HKL).

\subsection{Reformulation in terms of multiple kernel learning}

Following~\cite{rakoto,pontil-jmlr}, we can simply derive an equivalent formulation of \eq{milasso2}. Using  Cauchy-Schwarz inequality, we have that for all $\eta \in \rb^V$
such that $ \eta \geqslant 0$ and $ \sum_{v \in V}  d_v^2 \eta_v \leqslant 1$,

 \vspace*{-.4cm}

$$ \textstyle 
 ( \sum_{v \in V} d_v \| \beta_{\des(v)} \|  )^2  
 \leqslant  \sum_{v \in V}
\frac{ \| \beta_{\des(v)}\|^2}{ \eta_v}
= \sum_{w \in V} ( \sum_{v \in \anc(w) } \eta_v^{-1} ) \| \beta_w\|^2,
 $$

 \vspace*{-.2cm}

 with equality if and only if $\eta_v  = d_v^{-1} \| \beta_{\des(v)}\|  (\sum_{v \in V} d_v
 \| \beta_{\des(v) } \| )^{-1}$.
 We associate to the vector $\eta \in \rb^V$, the vector $\zeta \in \rb^V$ such that
$\forall w \in V$,
$\zeta_w^{-1}  =   \sum_{v \in \anc(w)} \eta_v^{-1}$. We use the natural convention that if $\eta_v$ is equal to zero, then $\zeta_w$ is equal to zero for all  descendants  $w$ of $v$. We let denote $H$ the set of allowed $\eta$ and $Z$ the set of all associated $\zeta$. The set $H$ and $Z$ are in bijection, and we can interchangeably use $\eta
\in H$ or the corresponding $\zeta(\eta) \in Z$.
Note that $Z$ is in general not convex (unless the DAG is a tree, see the Appendix), and if $\zeta \in Z$, then $\zeta_w \leqslant \zeta_{v}$ for all $w \in \des(v)$, i.e., weights of descendant kernels are smaller, which is consistent with the known fact that kernels should always be selected after all their ancestors.

 The problem in \eq{milasso2} is thus equivalent to

 \vspace*{-.4cm}

\BEQ
\label{eq:milasso2-eq}
 {
\min_{ \eta \in H}
\min_{   \beta\in \prod_{v\in V} \! \mathcal{F}_v} \textstyle
\frac{1}{n} \sum_{i=1}^n \ell(y_i, \sum_{v \in V} \langle \beta_v , \Phi_v(x_i) \rangle )
+ \frac{\lambda }{2}
\sum_{w \in V}  \zeta_w(\eta)^{-1} \|\beta_w\|^2}.
\EEQ

 \vspace*{-.4cm}

Using the change of variable $\tilde{\beta}_v = \beta_v \zeta_v^{-1/2}$ and $\tilde{\Phi}(x) = ( \zeta_v^{1/2} \Phi_v(x))_{v \in V}$, this implies that given the optimal $\eta$ (and associated $\zeta$),   $\beta$ corresponds to the solution of the regular supervised learning problem with kernel matrix $K = \sum_{w \in V}
\zeta_w K_w$, where $K_w$ is $n \times n$ the kernel matrix associated with kernel $k_w$. Moreover, the solution is then  $\beta_w = \zeta_w \sum_{i=1}^n \alpha_i \Phi_w(x_i)$, where $\alpha \in \rb^n$ are the dual parameters associated with the single kernel learning problem.

Thus, the solution is entirely determined by $\alpha \in \rb^n$ and $\eta \in \rb^V$ (and its corresponding $\zeta \in \rb^V$). More precisely, we have (see proof in the Appendix):
\begin{proposition} The pair
$(\alpha,\eta)$ is optimal for \eq{milasso2}, with $\forall w, \beta_w\! =\! \zeta_w \sum_{i=1}^n \alpha_i \Phi_w(x_i)$, if and only if  $(a)  $ given $\eta$,  $\alpha$ is optimal for the single kernel learning problem with kernel matrix 
 $K = \sum_{w\in V} \zeta_w(\eta)  K_w$, and   $(b) $ given $\alpha$, $\eta \in H $ maximizes $$
  \sum_{w \in V}  ( \sum_{v \in \anc(w) } \eta_v^{-1}  )^{-1}\alpha^\top K_w \alpha.$$ 
 \end{proposition}
  Moreover, the total duality gap can be upperbounded as the sum of the two separate duality gaps for the two optimization problems, which will be useful in \mysec{reduced} (see Appendix for more details).
   Note that in the case of ``flat'' regular  multiple kernel learning, where the DAG has no edges, we obtain back usual optimality conditions~\cite{rakoto,pontil-jmlr}.
 
 Following a common practice for convex sparsity problems~\cite{ng-sparsecoding}, we will try to solve a small problem where we assume we know the set of $v$ such that $\| \beta_{\des(v)}\|$ is equal to zero (\mysec{small-dual}). We then ``simply'' need to check that variables in that set may indeed be left out of the solution. In the next section, we show that this can be done in polynomial time although the number of kernels to consider leaving out is exponential (\mysec{reduced}).

\subsection{Conditions for global optimality of reduced problem}

\label{sec:reduced}
 We let denote $J$ the complement of the set of norms which are  set to zero. We thus consider the optimal solution $\beta$ of the reduced problem (on $J$), namely,
\BEQ
\label{eq:milasso-reduced}
\textstyle
\min_{ \beta_{J} \in \prod_{v\in J }\! \mathcal{F}_v} \textstyle
\frac{1}{n} \sum_{i=1}^n \ell(y_i,  \sum_{v \in J } \langle \beta_v , \Phi_v(x_i) \rangle )
+ \frac{\lambda }{2} \left(  \sum_{v \in V}  d_v \| \beta_{\des(v) \cap J}  \| \right)^2,
\EEQ
with optimal primal variables $\beta_J$, dual variables $\alpha$ and optimal pair $(\eta_J,\zeta_J)$.
We now consider necessary conditions and sufficient conditions for this solution (augmented with zeros for non active variables, i.e., variables in $J^c$) to
be optimal with respect to the full problem in \eq{milasso2}. We denote by $\delta = 
\sum_{v \in J} d_v \| \beta_{D(v) \cap J } \|$  the optimal value of the norm for the reduced problem.
\begin{proposition}[$N_{J}$]\hspace*{.1cm}
If the reduced solution is optimal for the full problem in \eq{milasso2} and all kernels in the extreme points of $J$ are active, then we have
$$  \max_{t \in \sou(J^c) }     {\alpha^\top K_t \alpha} /
{ d_t^2 } \leqslant
\delta^2 .$$
\end{proposition}

\vspace*{-.4cm}

\begin{proposition}[$S_{J,\varepsilon}$] \hspace*{.1cm}
If $ 
\max_{ t \in \sou(J^c)}   \textstyle
\sum_{w \in \des(t)}  \alpha^\top K_w \alpha / 
(\sum_{v \in \anc(w) \cap \des(t) } d_{v} )^2
 \leqslant  \delta^2 + \varepsilon/\lambda
$, then the total duality gap is less than $  \varepsilon$.
\end{proposition}
The proof is fairly technical and can be found in the Appendix; this result constitutes the main technical contribution of the paper: it essentially allows to solve a very large optimization problem over exponentially many dimensions in polynomial time.  
 
  The necessary condition $(N_J)$ does not cause any computational problems. However, the sufficient condition $(S_{J,\varepsilon})$ requires to sum over all descendants of the active kernels, which is impossible in practice (as shown in \mysec{simulations}, we consider $V$ of cardinal often greater than $10^{30}$). Here, we need to bring to bear the specific structure of the kernel $k$. In the context of directed grids we consider in this paper, if $d_v$ can also be decomposed as a product, then $\sum_{v \in \anc(w) \cap \des(t)} d_v $ is also factorized, and we can compute the sum over all $v \in \des(t)$ in linear time in $p$. Moreover we can cache the sums 
$
\sum_{w \in \des(t)}    K_w / 
(\sum_{v \in \anc(w) \cap \des(t) } d_{v} )^2 
$ in order to save running time.

\subsection{Dual optimization for reduced or small problems}

\label{sec:small-dual}
When kernels $k_v$, $v \in V$ have low-dimensional feature spaces, we may use a primal representation and solve the problem in \eq{milasso2} using generic optimization toolboxes adapted to conic constraints (see, e.g.,~\cite{boyd}). However, in order to reuse existing optimized supervised learning code and use  high-dimensional kernels, it is preferable to use a dual optimization. Namely, we use the same technique as~\cite{rakoto}:   we consider for $\zeta \in Z$, the function $B(\zeta)= 
\min_{   \beta\in \prod_{v\in V} \! \mathcal{F}_v} \textstyle
\frac{1}{n} \sum_{i=1}^n \ell(y_i, \sum_{v \in V} \langle \beta_v , \Phi_v(x_i) \rangle )
+ \frac{\lambda }{2}
\sum_{w \in V}  \zeta_w^{-1} \|\beta_w\|^2$, which is the optimal value of the single kernel learning problem with kernel matrix $\sum_{w \in V} \zeta_w K_w$. Solving \eq{milasso2-eq} is equivalent to minimizing $B(\zeta(\eta))$ with respect to $\eta \in H$.

 If a ridge (i.e., positive diagonal) is added to the kernel matrices, the function $B$ is differentiable. Moreover, the function $\eta \mapsto \zeta(\eta)$ is differentiable on $(\rb_+^\ast)^V$. Thus,
the function $\eta \mapsto 
 B[ \zeta(  (1-\varepsilon)  \eta+ \frac{\varepsilon}{|V|}  d^{-2}  )]$ , where $d^{-2}$ is the vector with elements $d_v^{-2}$, is differentiable if $\varepsilon>0$. We can then use the same projected gradient descent strategy as~\cite{rakoto} to minimize it. The overall complexity of the algorithm is then proportional to $O(|V| n^2)$---to form the kernel matrices---plus the complexity of solving a single kernel learning problem---typically between $O(n^2)$ and $O(n^3)$.

\subsection{Kernel search algorithm}

\label{sec:algorithm}
We are now ready to  present the detailed algorithm which extends the feature search algorithm of~\cite{ng-sparsecoding}. Note that the kernel matrices are never all needed explicitly, i.e., we only need them (a) explicitly to solve the small problems (but we need only a few of those) and (b)   implicitly to compute the sufficient condition
$(S_{J,\varepsilon})$, which requires to sum over all kernels, as shown in \mysec{reduced}.

\vspace*{-.2cm}

\BIT
\item \textbf{Input}: kernel matrices $K_v  \in \rb^{n \times n}$, $v \in V$, maximal gap $\varepsilon$, maximal $\#$ of kernels $Q$

\vspace*{-.15cm}

\item \textbf{Algorithm}

\vspace*{-.1cm}

\BNUM
\item Initialization: set $J = \sou(V)$, \\
\hspace*{1.93cm} compute $(\alpha,\eta)$ solutions of
\eq{milasso-reduced}, obtained using \mysec{small-dual}

\item while $(N_J)$ and $(S_{J,\varepsilon})$  are not satisfied and
$\#(V) \leqslant Q$
\BIT
\item 
If $(N_J)$ is not satisfied, add violating variables in $\sou(J^c)$ to $J$ \\
\hspace*{2.53cm} else, add violating variables in $\sou(J^c)$ of $(S_{J,\varepsilon})$ to $J$
\item Recompute   $(\alpha,\eta)$ optimal solutions of
\eq{milasso-reduced}

\EIT
 
\ENUM

\vspace*{-.2cm}

\item \textbf{Output}: $J$, $\alpha$, $\eta$
\EIT

\vspace*{-.2cm}

The previous algorithm will stop either when the duality gap is less than $\varepsilon$ or when the maximal number of kernels $Q$ has been reached. In practice, when the weights $d_v$ increase with the depth of $v$ in the DAG (which we use in simulations), the small duality gap generally occurs before we reach a problem larger than $Q$. Note that some of the iterations only increase the size of the active sets to check the sufficient condition for optimality; forgetting those does not change the solution, only the fact that we may actually know that we have an $\varepsilon$-optimal solution.

In the directed $p$-grid case, the total running time complexity is a function of the   number of observations $n$, and the number $R$ of selected kernels; with proper caching, we obtain the following complexity, assuming $O(n^3)$ for the single kernel learning problem, which is conservative: $O( n^3 R + n^2 R p^2 + n^2 R^2 p)$, which decomposes into solving $O(R)$ single kernel learning problems, caching $O(Rp)$ kernels, and computing $O(R^2p)$ quadratic forms for the sufficient conditions. 
Note that the kernel search algorithm is also an efficient algorithm for 
unstructured MKL.

\section{Consistency conditions}

\label{sec:consistency}

As said earlier, the sparsity pattern of the solution of \eq{milasso2} will be equal to its hull, and thus we can only hope to obtain consistency of the hull of the pattern, which we consider in this section.

For simplicity, we consider the case of finite dimensional Hilbert spaces (i.e., $\F_v = \rb^{f_v}$) and the square loss. We also hold fixed the vertex set of $V$, i.e., we assume that the total number of features is fixed, and we let $n$ tend to infinity and $\lambda = \lambda_n$ decrease with $n$.  

Following \cite{grouplasso}, we make the following assumptions on the underlying joint distribution of $(X,Y)$: (a) the joint covariance matrix $\boldsymbol\Sigma$ of $(\Phi(x_v))_{v \in V}$ (defined with appropriate blocks of size $f_v \times f_w$) is invertible, (b) $E(Y|X) = \sum_{w \in \boldsymbol W} \langle
\boldsymbol\beta_w,\Phi_w(x) \rangle$ with $\boldsymbol W \subset V$ and $\var(Y|X) = \boldsymbol\sigma^2 >0$ almost surely.
With these simple assumptions, we obtain (see proof in the Appendix):
\begin{proposition}[Sufficient condition]
If we have
$$\!\!\!\! \displaystyle \max_{t \in \sou(\boldsymbol W^c)} \textstyle \!\!\sum_{w \in \des(t)}  \!\! \frac{
\| \boldsymbol \Sigma_{w \boldsymbol W} \boldsymbol \Sigma_{\boldsymbol W \boldsymbol W  }^{-1} \Diag(d_v\| \boldsymbol \beta_{\des(v)} \|^{-1} )_{v \in \boldsymbol W } \boldsymbol \beta_{ \boldsymbol  W}\|^2}
{ ( \sum_{v \in \anc(w) \cap \des(t) } d_v)^2 }<1,$$
 then $\boldsymbol \beta$ and the hull of $\boldsymbol W$ are consistently estimated when $\lambda_n n^{1/2}\to \infty$ and $\lambda_n \to 0$.
\end{proposition}

\vspace*{-.3cm}

\begin{proposition}[Necessary condition]
If the $\boldsymbol \beta$ and the hull of $\boldsymbol W$ are consistently estimated for some sequence $\lambda_n$, then  
$$  \max_{t \in \sou(\boldsymbol  W^c)}  
\| \boldsymbol \Sigma_{w \boldsymbol  W} \boldsymbol \Sigma_{\boldsymbol  W   \boldsymbol  W  }^{-1} \Diag(d_v/\|  \boldsymbol \beta_{\des(v)} \| )_{v \in \boldsymbol  W } \boldsymbol \beta_{\boldsymbol  W }\|^2
/ d_t^2
 \leqslant 1.$$
 \end{proposition}
 Note that the last two propositions are not consequences of the similar results for flat MKL~\cite{grouplasso}, because the groups that we consider are overlapping.
 Moreover,  the last propositions show that we indeed can estimate the correct hull of the sparsity pattern if the sufficient condition is satisfied. In particular, if we can make the groups such that the between-group correlation is as small as possible, we can ensure correct hull selection. 
 Finally, it is worth noting that if the ratios $d_w / \max_{v \in \anc(w) } d_v$ tend to infinity slowly with $n$, then we always consistently estimate the depth of the hull, i.e., the optimal interaction complexity.
We are currently investigating extensions to the non parametric case~\cite{grouplasso}, in terms of pattern selection and universal consistency.

      \begin{figure}

\vspace*{-.45cm}

\begin{center}
\includegraphics[scale=.6475]{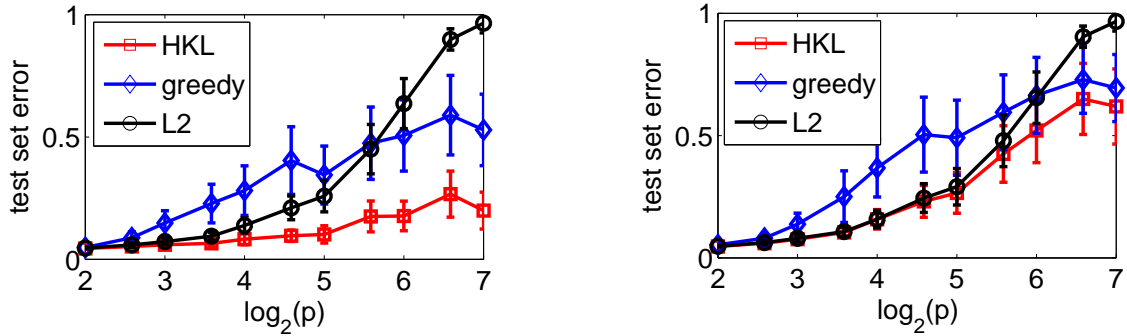} 

\vspace*{-.74cm}

\end{center}
\caption{Comparison on synthetic examples: mean squared error over 40 replications (with halved standard deviations). Left: non rotated data, right: rotated data. See text for details.
}
\label{fig:toy}
\end{figure}

   \section{Simulations}
   \label{sec:simulations}

   \textbf{Synthetic examples} \hspace*{.2cm}
   We generated regression data as follows: $n=1024$ samples of  $p \in [2^2, 2^7]$ variables were generated from a random covariance matrix, and the label $y \in \rb$ was sampled as a random sparse fourth order polynomial of the input variables (with constant number of monomials). We then compare the performance of our hierarchical multiple kernel learning method (HKL) with the polynomial kernel decomposition presented in \mysec{mkl} to other methods that use the same kernel and/or decomposition: (a) the greedy strategy of selecting basis kernels one after the other, a procedure similar to~\cite{bennett}, and (b) the regular polynomial kernel regularization with the full kernel (i.e., the sum of all basis kernels). In \myfig{toy}, we compare the two approaches on 40 replications in the following two situations: original data (left) and rotated data (right), i.e., after the input variables were transformed by a random rotation (in this situation, the generating polynomial is not sparse anymore). We can see that in situations where the underlying predictor function is sparse (left), HKL outperforms the two other methods when the total number of variables $p$ increases, while in the other situation where the best predictor is not sparse (right), it performs only slightly better: i.e., in non sparse problems, $\ell^1$-norms do not really help, but do help a lot when sparsity is expected.

\begin{table}

\vspace*{-.1cm}

\begin{center}
\hspace*{-1cm}
\begin{tabular}{|ccccc|ccccc|}
 \hline 
dataset & $n$ & $p$ & $k$ & $\#(V)$ &  L2 & greedy & lasso-$\alpha$ & MKL & HKL \\ 
\hline 
{     abalone} $\!\!\!\!$ &  $\!\!\!\!$   4177  $\!\!\!\!$ &  $\!\!\!\!$     10 $\!\!\!\!$  & $\!\!\!\!$    pol4 $\!\!\!\!$  &  $\!\!\!\!$ $\approx \!\! 10^{7}$  $\!\!\!\!$  &  $\!\!\!$ 44.2 $\!\!\pm\!\!$ 1.3 $\!\!\!$&  $\!\!\!$ 43.9 $\!\!\pm\!\!$ 1.4 $\!\!\!$&  $\!\!\!$ 47.9 $\!\!\pm\!\!$ 0.7 $\!\!\!$&  $\!\!\!$ 44.5 $\!\!\pm\!\!$ 1.1 $\!\!\!$& $\!\!\!$ \textbf{43.3 $\!\!\pm\!\!$ 1.0 $\!\!\!$}  \\ 
{     abalone} $\!\!\!\!$ &  $\!\!\!\!$   4177  $\!\!\!\!$ &  $\!\!\!\!$     10 $\!\!\!\!$  & $\!\!\!\!$     rbf $\!\!\!\!$  &  $\!\!\!\!$ $\approx \!\! 10^{10}$  $\!\!\!\!$  & $\!\!\!$ \textbf{43.0 $\!\!\pm\!\!$ 0.9 $\!\!\!$} &  $\!\!\!$ 45.0 $\!\!\pm\!\!$ 1.7 $\!\!\!$&  $\!\!\!$ 49.0 $\!\!\pm\!\!$ 1.7 $\!\!\!$&  $\!\!\!$ 43.7 $\!\!\pm\!\!$ 1.0 $\!\!\!$&  $\!\!\!$ 43.0 $\!\!\pm\!\!$ 1.1 $\!\!\!$ \\ 
 \hline 
{   bank-32fh} $\!\!\!\!$ &  $\!\!\!\!$   8192  $\!\!\!\!$ &  $\!\!\!\!$     32 $\!\!\!\!$  & $\!\!\!\!$    pol4 $\!\!\!\!$  &  $\!\!\!\!$ $\approx \!\! 10^{22}$  $\!\!\!\!$  &  $\!\!\!$ 40.1 $\!\!\pm\!\!$ 0.7 $\!\!\!$&  $\!\!\!$ 39.2 $\!\!\pm\!\!$ 0.8 $\!\!\!$&  $\!\!\!$ 41.3 $\!\!\pm\!\!$ 0.7 $\!\!\!$& $\!\!\!$ \textbf{38.7 $\!\!\pm\!\!$ 0.7 $\!\!\!$} &  $\!\!\!$ 38.9 $\!\!\pm\!\!$ 0.7 $\!\!\!$ \\ 
{   bank-32fh} $\!\!\!\!$ &  $\!\!\!\!$   8192  $\!\!\!\!$ &  $\!\!\!\!$     32 $\!\!\!\!$  & $\!\!\!\!$     rbf $\!\!\!\!$  &  $\!\!\!\!$ $\approx \!\! 10^{31}$  $\!\!\!\!$  &  $\!\!\!$ 39.0 $\!\!\pm\!\!$ 0.7 $\!\!\!$&  $\!\!\!$ 39.7 $\!\!\pm\!\!$ 0.7 $\!\!\!$&  $\!\!\!$ 66.1 $\!\!\pm\!\!$ 6.9 $\!\!\!$&  $\!\!\!$ 38.4 $\!\!\pm\!\!$ 0.7 $\!\!\!$& $\!\!\!$ \textbf{38.4 $\!\!\pm\!\!$ 0.7 $\!\!\!$}  \\ 
 \hline 
{   bank-32fm} $\!\!\!\!$ &  $\!\!\!\!$   8192  $\!\!\!\!$ &  $\!\!\!\!$     32 $\!\!\!\!$  & $\!\!\!\!$    pol4 $\!\!\!\!$  &  $\!\!\!\!$ $\approx \!\! 10^{22}$  $\!\!\!\!$  &  $\!\!\!$ 6.0 $\!\!\pm\!\!$ 0.1 $\!\!\!$& $\!\!\!$ \textbf{5.0 $\!\!\pm\!\!$ 0.2 $\!\!\!$} &  $\!\!\!$ 7.0 $\!\!\pm\!\!$ 0.2 $\!\!\!$&  $\!\!\!$ 6.1 $\!\!\pm\!\!$ 0.3 $\!\!\!$&  $\!\!\!$ 5.1 $\!\!\pm\!\!$ 0.1 $\!\!\!$ \\ 
{   bank-32fm} $\!\!\!\!$ &  $\!\!\!\!$   8192  $\!\!\!\!$ &  $\!\!\!\!$     32 $\!\!\!\!$  & $\!\!\!\!$     rbf $\!\!\!\!$  &  $\!\!\!\!$ $\approx \!\! 10^{31}$  $\!\!\!\!$  &  $\!\!\!$ 5.7 $\!\!\pm\!\!$ 0.2 $\!\!\!$&  $\!\!\!$ 5.8 $\!\!\pm\!\!$ 0.4 $\!\!\!$&  $\!\!\!$ 36.3 $\!\!\pm\!\!$ 4.1 $\!\!\!$&  $\!\!\!$ 5.9 $\!\!\pm\!\!$ 0.2 $\!\!\!$& $\!\!\!$ \textbf{4.6 $\!\!\pm\!\!$ 0.2 $\!\!\!$}  \\ 
 \hline 
{   bank-32nh} $\!\!\!\!$ &  $\!\!\!\!$   8192  $\!\!\!\!$ &  $\!\!\!\!$     32 $\!\!\!\!$  & $\!\!\!\!$    pol4 $\!\!\!\!$  &  $\!\!\!\!$ $\approx \!\! 10^{22}$  $\!\!\!\!$  &  $\!\!\!$ 44.3 $\!\!\pm\!\!$ 1.2 $\!\!\!$&  $\!\!\!$ 46.3 $\!\!\pm\!\!$ 1.4 $\!\!\!$&  $\!\!\!$ 45.8 $\!\!\pm\!\!$ 0.8 $\!\!\!$&  $\!\!\!$ 46.0 $\!\!\pm\!\!$ 1.2 $\!\!\!$& $\!\!\!$ \textbf{43.6 $\!\!\pm\!\!$ 1.1 $\!\!\!$}  \\ 
{   bank-32nh} $\!\!\!\!$ &  $\!\!\!\!$   8192  $\!\!\!\!$ &  $\!\!\!\!$     32 $\!\!\!\!$  & $\!\!\!\!$     rbf $\!\!\!\!$  &  $\!\!\!\!$ $\approx \!\! 10^{31}$  $\!\!\!\!$  &  $\!\!\!$ 44.3 $\!\!\pm\!\!$ 1.2 $\!\!\!$&  $\!\!\!$ 49.4 $\!\!\pm\!\!$ 1.6 $\!\!\!$&  $\!\!\!$ 93.0 $\!\!\pm\!\!$ 2.8 $\!\!\!$&  $\!\!\!$ 46.1 $\!\!\pm\!\!$ 1.1 $\!\!\!$& $\!\!\!$ \textbf{43.5 $\!\!\pm\!\!$ 1.0 $\!\!\!$}  \\ 
 \hline 
{   bank-32nm} $\!\!\!\!$ &  $\!\!\!\!$   8192  $\!\!\!\!$ &  $\!\!\!\!$     32 $\!\!\!\!$  & $\!\!\!\!$    pol4 $\!\!\!\!$  &  $\!\!\!\!$ $\approx \!\! 10^{22}$  $\!\!\!\!$  &  $\!\!\!$ 17.2 $\!\!\pm\!\!$ 0.6 $\!\!\!$&  $\!\!\!$ 18.2 $\!\!\pm\!\!$ 0.8 $\!\!\!$&  $\!\!\!$ 19.5 $\!\!\pm\!\!$ 0.4 $\!\!\!$&  $\!\!\!$ 21.0 $\!\!\pm\!\!$ 0.7 $\!\!\!$& $\!\!\!$ \textbf{16.8 $\!\!\pm\!\!$ 0.6 $\!\!\!$}  \\ 
{   bank-32nm} $\!\!\!\!$ &  $\!\!\!\!$   8192  $\!\!\!\!$ &  $\!\!\!\!$     32 $\!\!\!\!$  & $\!\!\!\!$     rbf $\!\!\!\!$  &  $\!\!\!\!$ $\approx \!\! 10^{31}$  $\!\!\!\!$  &  $\!\!\!$ 16.9 $\!\!\pm\!\!$ 0.6 $\!\!\!$&  $\!\!\!$ 21.0 $\!\!\pm\!\!$ 0.6 $\!\!\!$&  $\!\!\!$ 62.3 $\!\!\pm\!\!$ 2.5 $\!\!\!$&  $\!\!\!$ 20.9 $\!\!\pm\!\!$ 0.7 $\!\!\!$& $\!\!\!$ \textbf{16.4 $\!\!\pm\!\!$ 0.6 $\!\!\!$}  \\ 
 \hline 
{      boston} $\!\!\!\!$ &  $\!\!\!\!$    506  $\!\!\!\!$ &  $\!\!\!\!$     13 $\!\!\!\!$  & $\!\!\!\!$    pol4 $\!\!\!\!$  &  $\!\!\!\!$ $\approx \!\! 10^{9}$  $\!\!\!\!$  & $\!\!\!$ \textbf{17.1 $\!\!\pm\!\!$ 3.6 $\!\!\!$} &  $\!\!\!$ 24.7 $\!\!\pm\!\!$ 10.8 $\!\!\!$&  $\!\!\!$ 29.3 $\!\!\pm\!\!$ 2.3 $\!\!\!$&  $\!\!\!$ 22.2 $\!\!\pm\!\!$ 2.2 $\!\!\!$&  $\!\!\!$ 18.1 $\!\!\pm\!\!$ 3.8 $\!\!\!$ \\ 
{      boston} $\!\!\!\!$ &  $\!\!\!\!$    506  $\!\!\!\!$ &  $\!\!\!\!$     13 $\!\!\!\!$  & $\!\!\!\!$     rbf $\!\!\!\!$  &  $\!\!\!\!$ $\approx \!\! 10^{12}$  $\!\!\!\!$  & $\!\!\!$ \textbf{16.4 $\!\!\pm\!\!$ 4.0 $\!\!\!$} &  $\!\!\!$ 32.4 $\!\!\pm\!\!$ 8.2 $\!\!\!$&  $\!\!\!$ 29.4 $\!\!\pm\!\!$ 1.6 $\!\!\!$&  $\!\!\!$ 20.7 $\!\!\pm\!\!$ 2.1 $\!\!\!$&  $\!\!\!$ 17.1 $\!\!\pm\!\!$ 4.7 $\!\!\!$ \\ 
 \hline 
{pumadyn-32fh} $\!\!\!\!$ &  $\!\!\!\!$   8192  $\!\!\!\!$ &  $\!\!\!\!$     32 $\!\!\!\!$  & $\!\!\!\!$    pol4 $\!\!\!\!$  &  $\!\!\!\!$ $\approx \!\! 10^{22}$  $\!\!\!\!$  &  $\!\!\!$ 57.3 $\!\!\pm\!\!$ 0.7 $\!\!\!$&  $\!\!\!$ 56.4 $\!\!\pm\!\!$ 0.8 $\!\!\!$&  $\!\!\!$ 57.5 $\!\!\pm\!\!$ 0.4 $\!\!\!$& $\!\!\!$ \textbf{56.4 $\!\!\pm\!\!$ 0.7 $\!\!\!$} &  $\!\!\!$ 56.4 $\!\!\pm\!\!$ 0.8 $\!\!\!$ \\ 
{pumadyn-32fh} $\!\!\!\!$ &  $\!\!\!\!$   8192  $\!\!\!\!$ &  $\!\!\!\!$     32 $\!\!\!\!$  & $\!\!\!\!$     rbf $\!\!\!\!$  &  $\!\!\!\!$ $\approx \!\! 10^{31}$  $\!\!\!\!$  &  $\!\!\!$ 57.7 $\!\!\pm\!\!$ 0.6 $\!\!\!$&  $\!\!\!$ 72.2 $\!\!\pm\!\!$ 22.5 $\!\!\!$&  $\!\!\!$ 89.3 $\!\!\pm\!\!$ 2.0 $\!\!\!$&  $\!\!\!$ 56.5 $\!\!\pm\!\!$ 0.8 $\!\!\!$& $\!\!\!$ \textbf{55.7 $\!\!\pm\!\!$ 0.7 $\!\!\!$}  \\ 
 \hline 
{pumadyn-32fm} $\!\!\!\!$ &  $\!\!\!\!$   8192  $\!\!\!\!$ &  $\!\!\!\!$     32 $\!\!\!\!$  & $\!\!\!\!$    pol4 $\!\!\!\!$  &  $\!\!\!\!$ $\approx \!\! 10^{22}$  $\!\!\!\!$  &  $\!\!\!$ 6.9 $\!\!\pm\!\!$ 0.1 $\!\!\!$&  $\!\!\!$ 6.4 $\!\!\pm\!\!$ 1.6 $\!\!\!$&  $\!\!\!$ 7.5 $\!\!\pm\!\!$ 0.2 $\!\!\!$&  $\!\!\!$ 7.0 $\!\!\pm\!\!$ 0.1 $\!\!\!$& $\!\!\!$ \textbf{3.1 $\!\!\pm\!\!$ 0.0 $\!\!\!$}  \\ 
{pumadyn-32fm} $\!\!\!\!$ &  $\!\!\!\!$   8192  $\!\!\!\!$ &  $\!\!\!\!$     32 $\!\!\!\!$  & $\!\!\!\!$     rbf $\!\!\!\!$  &  $\!\!\!\!$ $\approx \!\! 10^{31}$  $\!\!\!\!$  &  $\!\!\!$ 5.0 $\!\!\pm\!\!$ 0.1 $\!\!\!$&  $\!\!\!$ 46.2 $\!\!\pm\!\!$ 51.6 $\!\!\!$&  $\!\!\!$ 44.7 $\!\!\pm\!\!$ 5.7 $\!\!\!$&  $\!\!\!$ 7.1 $\!\!\pm\!\!$ 0.1 $\!\!\!$& $\!\!\!$ \textbf{3.4 $\!\!\pm\!\!$ 0.0 $\!\!\!$}  \\ 
 \hline 
{pumadyn-32nh} $\!\!\!\!$ &  $\!\!\!\!$   8192  $\!\!\!\!$ &  $\!\!\!\!$     32 $\!\!\!\!$  & $\!\!\!\!$    pol4 $\!\!\!\!$  &  $\!\!\!\!$ $\approx \!\! 10^{22}$  $\!\!\!\!$  &  $\!\!\!$ 84.2 $\!\!\pm\!\!$ 1.3 $\!\!\!$&  $\!\!\!$ 73.3 $\!\!\pm\!\!$ 25.4 $\!\!\!$&  $\!\!\!$ 84.8 $\!\!\pm\!\!$ 0.5 $\!\!\!$&  $\!\!\!$ 83.6 $\!\!\pm\!\!$ 1.3 $\!\!\!$& $\!\!\!$ \textbf{36.7 $\!\!\pm\!\!$ 0.4 $\!\!\!$}  \\ 
{pumadyn-32nh} $\!\!\!\!$ &  $\!\!\!\!$   8192  $\!\!\!\!$ &  $\!\!\!\!$     32 $\!\!\!\!$  & $\!\!\!\!$     rbf $\!\!\!\!$  &  $\!\!\!\!$ $\approx \!\! 10^{31}$  $\!\!\!\!$  &  $\!\!\!$ 56.5 $\!\!\pm\!\!$ 1.1 $\!\!\!$&  $\!\!\!$ 81.3 $\!\!\pm\!\!$ 25.0 $\!\!\!$&  $\!\!\!$ 98.1 $\!\!\pm\!\!$ 0.7 $\!\!\!$&  $\!\!\!$ 83.7 $\!\!\pm\!\!$ 1.3 $\!\!\!$& $\!\!\!$ \textbf{35.5 $\!\!\pm\!\!$ 0.5 $\!\!\!$}  \\ 
 \hline 
{pumadyn-32nm} $\!\!\!\!$ &  $\!\!\!\!$   8192  $\!\!\!\!$ &  $\!\!\!\!$     32 $\!\!\!\!$  & $\!\!\!\!$    pol4 $\!\!\!\!$  &  $\!\!\!\!$ $\approx \!\! 10^{22}$  $\!\!\!\!$  &  $\!\!\!$ 60.1 $\!\!\pm\!\!$ 1.9 $\!\!\!$&  $\!\!\!$ 69.9 $\!\!\pm\!\!$ 32.8 $\!\!\!$&  $\!\!\!$ 78.5 $\!\!\pm\!\!$ 1.1 $\!\!\!$&  $\!\!\!$ 77.5 $\!\!\pm\!\!$ 0.9 $\!\!\!$& $\!\!\!$ \textbf{5.5 $\!\!\pm\!\!$ 0.1 $\!\!\!$}  \\ 
{pumadyn-32nm} $\!\!\!\!$ &  $\!\!\!\!$   8192  $\!\!\!\!$ &  $\!\!\!\!$     32 $\!\!\!\!$  & $\!\!\!\!$     rbf $\!\!\!\!$  &  $\!\!\!\!$ $\approx \!\! 10^{31}$  $\!\!\!\!$  &  $\!\!\!$ 15.7 $\!\!\pm\!\!$ 0.4 $\!\!\!$&  $\!\!\!$ 67.3 $\!\!\pm\!\!$ 42.4 $\!\!\!$&  $\!\!\!$ 95.9 $\!\!\pm\!\!$ 1.9 $\!\!\!$&  $\!\!\!$ 77.6 $\!\!\pm\!\!$ 0.9 $\!\!\!$& $\!\!\!$ \textbf{7.2 $\!\!\pm\!\!$ 0.1 $\!\!\!$}  \\ 
 \hline 

\end{tabular}
\end{center}

\vspace*{-.4cm}

\caption{Mean squared errors (multiplied by 100) on UCI   regression datasets, normalized so that the total variance to explain is 100. See text for details. }
\label{tab:reg}

\vspace*{-.15cm}

\end{table}
  
  \textbf{UCI datasets} \hspace*{.2cm}
For regression datasets, we compare HKL with polynomial (degree 4) and Gaussian-RBF kernels (each dimension decomposed into 9 kernels) to the following approaches with the same kernel: regular Hilbertian regularization (L2), same greedy approach as earlier (greedy), regularization by the $\ell^1$-norm directly on the vector $\alpha$, a strategy which is sometimes used in the context of sparse kernel learning~\cite{roth} but does not use the Hilbertian structure of the kernel (lasso-$\alpha$), multiple kernel learning with the 	$p$ kernels obtained by summing all kernels associated with a single variable, a strategy suggested by~\cite{skm} (MKL). For all methods, the kernels were held fixed, while in Table~\ref{tab:reg}, we report the performance for the best regularization parameters obtained by 10 random half splits. 

We can see from Table~\ref{tab:reg}, that HKL outperforms other methods, in particular for the datasets bank-32nm, bank-32nh, pumadyn-32nm, pumadyn-32nh, which are datasets dedicated to non linear regression. Note also, that we efficiently explore DAGs with very large numbers of vertices  $\#(V)$.

\begin{table}

\vspace*{-.1cm}

\begin{center}
 \begin{tabular}{|ccccc|ccc|}
 \hline 
dataset & $n$ & $p$ & $k$ & $\#(V)$ & L2 & greedy  & HKL \\ 
 \hline 
{   mushrooms}  $\!\!\!\!$ &  $\!\!\!\!$   1024  $\!\!\!\!$ &  $\!\!\!\!$    117  $\!\!\!\!$ &  $\!\!\!\!$   pol4  $\!\!\!\!$ & $\!\!\!\!$ $\approx \!\! 10^{82}$ $\!\!\!\!$ &  $\!\!\!$ 0.4 $\!\!\pm\!\!$ 0.4 $\!\!\!$& $\!\!\!$ \textbf{0.1 $\!\!\pm\!\!$ 0.1 $\!\!\!$} &  $\!\!\!$ 0.1 $\!\!\pm\!\!$ 0.2 $\!\!\!$ \\ 
{   mushrooms}  $\!\!\!\!$ &  $\!\!\!\!$   1024  $\!\!\!\!$ &  $\!\!\!\!$    117  $\!\!\!\!$ &  $\!\!\!\!$    rbf  $\!\!\!\!$ & $\!\!\!\!$ $\approx \!\! 10^{112}$ $\!\!\!\!$ & $\!\!\!$ \textbf{0.1 $\!\!\pm\!\!$ 0.2 $\!\!\!$} &  $\!\!\!$ 0.1 $\!\!\pm\!\!$ 0.2 $\!\!\!$&  $\!\!\!$ 0.1 $\!\!\pm\!\!$ 0.2 $\!\!\!$ \\ 
 \hline 
{    ringnorm}  $\!\!\!\!$ &  $\!\!\!\!$   1024  $\!\!\!\!$ &  $\!\!\!\!$     20  $\!\!\!\!$ &  $\!\!\!\!$   pol4  $\!\!\!\!$ & $\!\!\!\!$ $\approx \!\! 10^{14}$ $\!\!\!\!$ &  $\!\!\!$ 3.8 $\!\!\pm\!\!$ 1.1 $\!\!\!$&  $\!\!\!$ 5.9 $\!\!\pm\!\!$ 1.3 $\!\!\!$& $\!\!\!$ \textbf{2.0 $\!\!\pm\!\!$ 0.3 $\!\!\!$}  \\ 
{    ringnorm}  $\!\!\!\!$ &  $\!\!\!\!$   1024  $\!\!\!\!$ &  $\!\!\!\!$     20  $\!\!\!\!$ &  $\!\!\!\!$    rbf  $\!\!\!\!$ & $\!\!\!\!$ $\approx \!\! 10^{19}$ $\!\!\!\!$ & $\!\!\!$ \textbf{1.2 $\!\!\pm\!\!$ 0.4 $\!\!\!$} &  $\!\!\!$ 2.4 $\!\!\pm\!\!$ 0.5 $\!\!\!$&  $\!\!\!$ 1.6 $\!\!\pm\!\!$ 0.4 $\!\!\!$ \\ 
 \hline 
{    spambase}  $\!\!\!\!$ &  $\!\!\!\!$   1024  $\!\!\!\!$ &  $\!\!\!\!$     57  $\!\!\!\!$ &  $\!\!\!\!$   pol4  $\!\!\!\!$ & $\!\!\!\!$ $\approx \!\! 10^{40}$ $\!\!\!\!$ &  $\!\!\!$ 8.3 $\!\!\pm\!\!$ 1.0 $\!\!\!$&  $\!\!\!$ 9.7 $\!\!\pm\!\!$ 1.8 $\!\!\!$& $\!\!\!$ \textbf{8.1 $\!\!\pm\!\!$ 0.7 $\!\!\!$}  \\ 
{    spambase}  $\!\!\!\!$ &  $\!\!\!\!$   1024  $\!\!\!\!$ &  $\!\!\!\!$     57  $\!\!\!\!$ &  $\!\!\!\!$    rbf  $\!\!\!\!$ & $\!\!\!\!$ $\approx \!\! 10^{54}$ $\!\!\!\!$ &  $\!\!\!$ 9.4 $\!\!\pm\!\!$ 1.3 $\!\!\!$&  $\!\!\!$ 10.6 $\!\!\pm\!\!$ 1.7 $\!\!\!$& $\!\!\!$ \textbf{8.4 $\!\!\pm\!\!$ 1.0 $\!\!\!$}  \\ 
 \hline 
{     twonorm}  $\!\!\!\!$ &  $\!\!\!\!$   1024  $\!\!\!\!$ &  $\!\!\!\!$     20  $\!\!\!\!$ &  $\!\!\!\!$   pol4  $\!\!\!\!$ & $\!\!\!\!$ $\approx \!\! 10^{14}$ $\!\!\!\!$ & $\!\!\!$ \textbf{2.9 $\!\!\pm\!\!$ 0.5 $\!\!\!$} &  $\!\!\!$ 4.7 $\!\!\pm\!\!$ 0.5 $\!\!\!$&  $\!\!\!$ 3.2 $\!\!\pm\!\!$ 0.6 $\!\!\!$ \\ 
{     twonorm}  $\!\!\!\!$ &  $\!\!\!\!$   1024  $\!\!\!\!$ &  $\!\!\!\!$     20  $\!\!\!\!$ &  $\!\!\!\!$    rbf  $\!\!\!\!$ & $\!\!\!\!$ $\approx \!\! 10^{19}$ $\!\!\!\!$ & $\!\!\!$ \textbf{2.8 $\!\!\pm\!\!$ 0.6 $\!\!\!$} &  $\!\!\!$ 5.1 $\!\!\pm\!\!$ 0.7 $\!\!\!$&  $\!\!\!$ 3.2 $\!\!\pm\!\!$ 0.6 $\!\!\!$ \\ 
 \hline 
{     magic04}  $\!\!\!\!$ &  $\!\!\!\!$   1024  $\!\!\!\!$ &  $\!\!\!\!$     10  $\!\!\!\!$ &  $\!\!\!\!$   pol4  $\!\!\!\!$ & $\!\!\!\!$ $\approx \!\! 10^{7}$ $\!\!\!\!$ &  $\!\!\!$ 15.9 $\!\!\pm\!\!$ 1.0 $\!\!\!$&  $\!\!\!$ 16.0 $\!\!\pm\!\!$ 1.6 $\!\!\!$& $\!\!\!$ \textbf{15.6 $\!\!\pm\!\!$ 0.8 $\!\!\!$}  \\ 
{     magic04}  $\!\!\!\!$ &  $\!\!\!\!$   1024  $\!\!\!\!$ &  $\!\!\!\!$     10  $\!\!\!\!$ &  $\!\!\!\!$    rbf  $\!\!\!\!$ & $\!\!\!\!$ $\approx \!\! 10^{10}$ $\!\!\!\!$ &  $\!\!\!$ 15.7 $\!\!\pm\!\!$ 0.9 $\!\!\!$&  $\!\!\!$ 17.7 $\!\!\pm\!\!$ 1.3 $\!\!\!$& $\!\!\!$ \textbf{15.6 $\!\!\pm\!\!$ 0.9 $\!\!\!$}  \\ 
 \hline 
\end{tabular}
\end{center}

\vspace*{-.35cm}

\caption{Error rates (multiplied by 100) on UCI binary classification datasets. See text for details. }
\label{tab:classif}
\end{table}

For binary classification datasets, we compare HKL (with the logistic loss) to two other methods (L2, greedy) in Table~\ref{tab:classif}. For some datasets (e.g., spambase), HKL works better, but for some others, in particular when  the generating problem is known to be non sparse (ringnorm, twonorm), it performs slightly worse than other approaches.

\section{Conclusion}

  We have shown how to perform hierarchical multiple kernel learning (HKL) in polynomial time in the number of selected kernels. This framework may be applied to many positive definite kernels and  we have focused on polynomial and Gaussian kernels used for nonlinear variable selection.
  In particular,  this paper shows that trying to use $\ell^1$-type penalties may be advantageous inside the feature space.  We are currently investigating applications to other kernels, such as the pyramid match kernel~\cite{grauman}, string kernels, and graph kernels~\cite{Cristianini2004}.

  \vspace*{-.25cm}

\appendix

\section{Optimization results}

In this first section, we give proofs of all results related to the optimization problems. We first recall precisely how we obtained the relationships between $\eta$ and $\zeta$.
Using  Cauchy-Schwarz inequality, we know that for all $\eta \in \rb^V$
such that $ \eta \geqslant 0$ and $ \sum_{v \in V}  d_v^2 \eta_v \leqslant 1$,
$$
\left( \sum_{v \in V} d_v \| \beta_{\des(v)} \| \right)^2 \!\!\!
=\! \left( \sum_{v \in V} ( d_v \eta_v^{1/2})  \frac{ \| \beta_{\des(v)}\|}{ \eta_v^{1/2}} \right)^2 \!\!\! $$
$$
\leqslant  \! \sum_{v \in V} d_v^2 \eta_v \times  \sum_{v \in V}
\frac{ \| \beta_{\des(v)}\|^2}{ \eta_v}
\leqslant \sum_{w \in V}\! \left( \sum_{v \in \anc(w) } \eta_v^{-1} \right) \| \beta_w\|^2,
 $$
 with equality if and only if $\eta_v  = d_v^{-1} \| \beta_{\des(v)}\|  (\sum_{v \in V} d_v
 \| \beta_{\des(v) } \| )^{-1}$.

\subsection{Set of weights for trees}

When the DAG is a tree (i.e., when each vertex has at most one parent), then, without loss of generality we may consider that only one vertex has no parent (the root $r$) while all others $w$ have exactly one parent $\pi(w)$. In this situation, we have for all $v \neq r$,
$\zeta_{\pi(v)}^{-1} - \zeta_v^{-1} = - \eta_{\pi(v)}^{-1}$. Moreover, for all leaves $v$, $\zeta_v = \eta_v$. This implies that the constraint $\eta\geqslant 0$ is equivalent to
$\zeta \geqslant 0$ and for all $v \neq r$,  $\zeta_{\pi(v)} \geqslant  \zeta_v$. The final constraint $\sum_{v \in V} \eta_v d_v^2 \leqslant 1$, may then be written as:
$$
\sum_{v \neq r } d_v^2 \frac{1}{ \zeta_v^{-1} - \zeta_{\pi(v)}^{-1}}  + \sum_{v \mbox{ leaf} } \zeta_v d_v^2 \leqslant 1,$$
that is,
$$
\sum_{v \neq r } d_v^2 \left( \zeta_v + \frac{ \zeta_v^2 }{ \zeta_{\pi(v)} -\zeta_v  }  \right) + \sum_{v \mbox{ leaf} } \zeta_v d_v^2 \leqslant 1,$$
which is clearly convex~\cite{boyd}.
When the DAG is not a tree, we conjecture that the set $Z$ is not convex.

\subsection{Fenchel conjugates}

Following~\cite{bach_thibaux, sonnenburg}, in order to derive optimality conditions for all losses, we need to introduce Fenchel conjugates.
Let $\psi_i:\rb \mapsto \rb$, be the Fenchel conjugate~\cite{boyd} of the convex function $\varphi_i : a \mapsto \ell(y_i,a )$, defined
 as 
 $$\psi_i(b) = \max_{a \in \rb} ab - \varphi_i(a). $$
  The function
 $\psi_i$ is  always convex and, because we have assumed that $\varphi_i$ is convex and continuous, we can represent $\varphi_i$ as the Fenchel conjugate of $\psi_i$, i.e., for all $a \in \rb$,
 $$
 \varphi_i(a) = \max_{b \in \rb} ab - \psi_i(b).
 $$

In particular, we have for the following standard examples:
\BIT
\item
for \emph{least-squares regression}, we have
 $\varphi_i(a)= \frac{1}{2}(y_i -a)^2$ and $\psi_i(b) = \frac{1}{2} b^2 + by_i$,
\item  for \emph{logistic regression}, we have
 $\varphi_i(a)=\log(1+\exp(-y_i a_i))$, where $y_i \in \{-1,1\}$,
 and $\psi_i(b) = (1+b y_i) \log(1+b y_i) -b y_i \log(-b y_i)$
if $b y_i \in [-1,0]$, $+ \infty$ otherwise.
\item for \emph{support vector machine classification}, we have
 $\varphi_i(a)=\max(0,1-y_i a)$, where $y_i \in \{-1,1\}$,
 and $\psi_i(b) =y_i b$
if $b y_i \in [-1,0]$, $+ \infty$ otherwise.

\EIT

\subsection{Preliminary propositions}

We first recall the duality result for  the regular $\ell^2$-norm kernel learning problem:
\begin{proposition}
For all nonnegative $\zeta \in \rb^V$, the dual of the optimization problem  
$$\min_{   \beta\in \prod_{v\in V} \! \mathcal{F}_v} \textstyle
\frac{1}{n} \sum_{i=1}^n \ell(y_i, \sum_{v \in V} \langle \beta_v , \Phi_v(x_i) \rangle )
+ \frac{\lambda }{2}
\sum_{w \in V}  \zeta_w^{-1} \|\beta_w\|^2$$ is
$$
\max_{\alpha \in \rb^n}
- \frac{1}{n} \sum_{i=1}^n \psi_i(-n\lambda \alpha_i)  
 - \frac{\lambda}{2}  \alpha^\top \left( \sum_{w \in V} \zeta_w K_w \right) \alpha,
$$
and the optimal $\beta$ can be found from an optimal $\alpha$ as 
$\beta_w = \sum_{i=1}^n \alpha_i \Phi_w(x_i)$.
\end{proposition}
\begin{proof}
We introduce auxiliary variables $u_i =  \sum_{v \in V} \langle \beta_v , \Phi_v(x_i) \rangle$ and consider the Lagrangian:
$$\mathcal{L} = \frac{1}{n} \sum_{i=1}^n \varphi_i(u_i)
+\frac{\lambda}{2} \sum_{w \in V}  \zeta_w^{-1} \|\beta_w\|^2  + \lambda \sum_{i=1}^n
 \alpha_i  ( u_i - \sum_{v \in V} \langle \beta_v , \Phi_v(x_i) \rangle)$$
Minimizing with respect to the primal variables $u,\beta$, we get the dual problem.
\end{proof}

We will   use the following simple result, which implies that each component $\zeta_w(\eta)$ is a concave function of   $\eta$:
\begin{lemma}
The minimum of $\sum_{j=1}^m a_j x_j^2$ subject to $\sum_{j=1}^m x_j=1$ is equal to
$\left( \sum_{j=1}^m a_i^{-1} \right)^{-1}$ and is attained at
$x_i = a_i^{-1}  \left( \sum_{j=1}^m a_i^{-1} \right)^{-1}$.
\end{lemma}
 
 The following proposition derives the dual of the problem in $\eta$:

 \begin{proposition} Let $L = \{ \kappa \in \rb^{V \times V}, \forall w \in V, \sum_{v \in \anc(w)} \kappa_{vw}=1\}$. The following optimization problems are dual to each other, and there is no duality gap : 
 $$ \displaystyle
\min_{\kappa \in L}
\max_{v \in V} d_v^{-2} \sum_{w \in \des(v)}\kappa_{vw}^2 \alpha^\top K_w
\alpha$$
$$  
\max_{\eta \in H} \sum_{w \in V}
\alpha^\top \zeta_w(\eta) K_w \alpha
 .$$
 \end{proposition}
 \begin{proof}
We have the Lagrangian
 $$\mathcal{L} 
 = \delta^2 + \sum_{v \in V} \eta_v \left( 
 \sum_{w \in \des(v)} \kappa_{vw}^2 \alpha^\top K_w \alpha  - \delta^2 d_v^2\right),$$
 which can be minimized in closed form with respect to $\delta^2$ and $\kappa \in L$, and leads to (using Lemma~1):
 $$\min_{\kappa \in L}
 \max_{v \in V}
d_v^{-2}  \sum_{w \in \des(v) } \kappa_{vw}^2 \alpha^\top K_w \alpha = \max_{\eta} \alpha^\top  \left(
\sum_{w \in V}  \zeta_w(\eta) K_w \right) \alpha.
$$
 \end{proof}

\subsection{Duality gaps}
We consider the following function of $\eta \in H$ and $\alpha \in \rb^n$:
$$F(\eta,\alpha) = 
- \frac{1}{n} \sum_{i=1}^n \psi_i(-n\lambda \alpha_i)  
 - \frac{\lambda}{2}  \alpha^\top  \sum_{w \in V} \zeta_w(\eta)K_w \alpha.
$$
This function is convex in $\eta$ (because of Lemma 1) and concave in $\alpha$, standard arguments (e.g., primal and dual strict feasibilities) show that there is no duality gap to the variational problems:
$$
\inf_{\eta \in H } \sup_{\alpha \in \rb^n} F(\eta,\alpha)
 =   \sup_{\alpha \in \rb^n} \inf_{\eta \in H } F(\eta,\alpha).
$$
We can decompose the duality gap, given a pair $(\eta,\alpha)$ as
\BEAS
& & \sup_{\alpha' \in \rb^n} F(\eta,\alpha')  -  \inf_{\eta' \in H } F(\eta',\alpha)  \\
&\!\!  = \!\! &   \min_{\beta}  \left\{
 \frac{1}{n} \sum_{i=1}^n \ell(y_i, \sum_{v \in V} \langle \beta_v , \Phi_v(x_i) \rangle )
+ \frac{\lambda }{2}
\sum_{w \in V}  \zeta_w(\eta)^{-1} \|\beta_w\|^2 \right\}  \\
& & -  \inf_{\eta' \in H } F(\eta',\alpha)  
\\
&\!\! \leqslant  \!\!&  
 \frac{1}{n} \sum_{i=1}^n \ell(y_i, \sum_{w \in V}\zeta_w(\eta) (K_w \alpha)_{i} )
+ \frac{\lambda }{2}
\sum_{w \in V}  \zeta_w \alpha^\top K_w \alpha + \frac{1}{n} \sum_{i=1}^n \psi_i(-n\lambda \alpha_i) \\
& & +  \sup_{\eta' \in H }  \frac{\lambda}{2}  \alpha^\top  \sum_{w \in V} \zeta_w(\eta') \alpha \\
&\!\! =  \!\!&  
 \frac{1}{n} \sum_{i=1}^n \ell(y_i, \sum_{w \in V}\zeta_w(\eta) (K_w \alpha)_{i} )
+ \frac{1}{n} \sum_{i=1}^n \psi_i(-n\lambda \alpha_i)  +  {\lambda }
\sum_{w \in V}  \zeta_w(\eta) \alpha^\top K_w \alpha  \\
& & +  \sup_{\eta' \in H }  \frac{\lambda}{2}  \alpha^\top  \sum_{w \in V} \zeta_w(\eta') \alpha
- \frac{\lambda }{2}
\sum_{w \in V}  \zeta_w(\eta)  \alpha^\top K_w \alpha .
\EEAS
We thus get the desired upper bound from which proposition 1 (of the main paper) follows, as well as the upper bound on the duality gap.

\subsection{Necessary and sufficient conditions - truncated problem}
We assume that we know the optimal solution of a truncated problem
where the entire set of decendants of some nodes have been removed. We let denote $J$ the hull of the set of active variables.
We now consider necessary conditions and sufficient conditions for this solution to
be optimal with respect to the full problem.  
This will lead to Proposition~2 and~3 of the main paper.

We first use Proposition~2 of the Appendix, to get a set of $\kappa_{vw}$ for $(v,w)\in J$ for the reduced problem; the goal here is to get necessary conditions by relaxing the dual problem defining $\kappa \in L$ and find an approximate solution, while for the sufficient condition, any candidate leads to a sufficient condition. It turns out that we will use the solution of the relaxed solution required for the necessary condition for the sufficient condition.

If we assume that all variables in $J$ are indeed active, then any optimal $\kappa \in L$ must be such that $\kappa_{vw} = 0$ if $v \in J$ and $w \in J^c$. We then let free 
$\kappa_{vw}$ for $v,w$ in $J$. Our goal is to find good candidates for those free dual parameters.

We first derive necessary conditions by lowerbounding the sums by maxima:
$$\max_{v \in V \cap J^c}
d_v^{-2}  \sum_{w \in \des(v) } \kappa_{vw}^2 \alpha^\top K_w \alpha
\geqslant 
\max_{v \in V \cap J^c}
d_v^{-2}  \max_{w \in \des(v) } \kappa_{vw}^2 \alpha^\top K_w \alpha,
$$
which can be minimized in closed form with respect to $\kappa$ leading to
$$\kappa_{vw} = d_v ( \sum_{v' \in A(w) \cap J^c} d_{v'} )^{-1}$$ and to the lower bound
\BEQ
 {\min_{\kappa \in L} \max_{v \in V \cap J^c}
d_v^{-2}  \sum_{w \in \des(v) } \kappa_{vw}^2 \alpha^\top K_w \alpha
\geqslant 
\max_{w \in J^c } \frac{\alpha^\top K_w \alpha}
{ (\sum_{v \in A(w) \cap J^c} d_{v})^2}}.
\EEQ

For sufficient conditions, we simply take the value obtained before for $\kappa$, which leads to
$$
\min_{\kappa \in L} \max_{v \in V \cap I^c}
d_v^{-2}  \sum_{w \in \des(v) } \kappa_{vw}^2 \alpha^\top K_w \alpha
\leqslant \max_{v \in V \cap J^c}
\sum_{w \in \des(v) } \frac{\alpha^\top K_w \alpha}
{ (\sum_{v \in A(w) \cap J^c} d_{v})^2}$$
$$
=
\max_{v \in  \sou(J^c) }
\sum_{w \in \des(v) } \frac{\alpha^\top K_w \alpha}
{ (\sum_{v \in A(w) \cap J^c} d_{v})^2}.
$$

 We have moreover
 $$
   \sum_{v \in A(w) } d_{v}   \geqslant \sum_{v \in A(w) \cap J^c} d_{v} \geqslant
   \sum_{v \in A(w) \cap D(t) } d_{v},
 $$
 leading to the desired upper bound
\BEQ
 { \min_{\kappa \in L} \max_{v \in V \cap J^c}
d_v^{-2}  \sum_{w \in \des(v) } \kappa_{vw}^2 \alpha^\top K_w \alpha
\leqslant 
\max_{ t \in \sou(J^c)}
\sum_{w \in D(t)} \frac{ \alpha^\top K_w \alpha}{
(\sum_{v \in A(w) \cap D(t) } d_{v} )^2
}
}.
\EEQ

\subsection{Optimality conditions for the primal formulation}
We know derive optimality conditions for the problem in the paper, which we will need in \mysec{consistency-proof}, i.e.:

$${
\min_{ \beta \in \prod_{v\in V}\! \mathcal{F}_v} \textstyle
\frac{1}{n} \sum_{i=1}^n \varphi_i(\sum_{v \in V} \langle \beta_v , \Phi_v(x_i) \rangle )
+ \frac{\lambda }{2} \left(  \sum_{v \in V}  d_v \| \beta_{\des(v)}  \| \right)^2}.
$$
Let $\beta\in \rb^V$, with $J$ being the   hull of the active variables. The directional derivative in the direction $\Delta \in \rb^V$ is equal to
$$
\frac{1}{n} \sum_{i=1}^n \sum_{w\in V} \varphi_i'(\sum_{v \in J} \langle \beta_v , \Phi_v(x_i) \rangle )
\Phi_w(x_i)^\top \Delta_w  $$
$$ + \lambda 
\left(  \sum_{v \in J}  d_v \| \beta_{\des(v)}  \| \right)
 \left(  \sum_{v \in J} d_v  \frac{ \beta_{\des(v) \cap J}}{\| \beta_{\des(v) \cap J}\| } ^\top \Delta_v   + \sum_{v \in J^c} d_v \| \Delta_{\des(v)} \|\right)
$$
and thus $\beta$ if optimal if and ony if, we have, with $\delta = \sum_{v \in J}  d_v \| \beta_{\des(v) \cap J }  \|  $:
$$
\forall w \in J, \
\frac{1}{n} \sum_{i=1}^n \varphi_i'(\sum_{v \in J} \langle \beta_v , \Phi_v(x_i) \rangle )
\Phi_w(x_i) + \lambda  \delta
 \left(
 \sum_{v \in \anc(w)}   \frac{  d_v }{\| \beta_{\des(v) \cap J}\| }  \right) \beta_w = 0
$$
 $$
 \forall \Delta_{J^c} \in \rb^{J^c}, \ 
 \frac{1}{n} \sum_{i=1}^n   \sum_{w\in J^c} \varphi_i'(\sum_{v \in J} \langle \beta_v , \Phi_v(x_i) \rangle )
\Phi_w(x_i)^\top \Delta_w + \lambda 
 \delta
 \left(  \sum_{v \in J^c} d_v \| \Delta_{\des(v)} \|\right) \geqslant 0.
 $$

Note that when regularizing by  $ {\lambda } \sum_{v \in V}  d_v \| \beta_{\des(v)}  \| $ instead of $\frac{\lambda }{2} \left(  \sum_{v \in V}  d_v \| \beta_{\des(v)}  \| \right)^2$, we have the same optimality condition with $\delta=1$.

\section{Consistency conditions}

We assume that we are in the finite dimensional setting (i.e., each $\mathcal{F}_v$ has finite dimensions $f_v$) with the square loss. For $w \in V$, we let denote $X_{w} \in \rb^{n \times f_w}$ the matrix whose $n$-th row is $\Phi_w(x_i)$. We let denote $\boldsymbol\Sigma_{vw} \in \rb^{f_v \times f_w}$ the population covariance between $\Phi_v(x)$ and $\Phi_w(x)$. The full covariance matrix, defined from the blocks $\boldsymbol \Sigma_{vw}$ is assumed invertible. With these assumptions, we can follow the approach of \cite{zou,fu,yuanlin} : that is, if $\lambda_n$ tends to zero faster than $n^{-1/2}$, then the estimate $\hat{\beta}$ converges in probability to the generating $\boldsymbol\beta$, and we have the expansion $\hat{\beta} = \boldsymbol\beta + \lambda_n \hat{\gamma}$ where $\hat{\gamma}$ is the solution of the following optimization problem, with
$\delta = \sum_{v \in  \boldsymbol W }  d_v \| \boldsymbol \beta_{\des(v)}  \|  $:
$$  
\min_{\gamma \in \prod_w \rb^{f_w} } \frac{1}{2} \gamma ^\top \boldsymbol\Sigma \gamma  +  \delta 
\sum_{v \in  \boldsymbol W } d_v  \frac{ \boldsymbol\beta_{\des(v) \cap  \boldsymbol W }}{\| \boldsymbol\beta_{\des(v) \cap  \boldsymbol W }\| } ^\top \gamma_v   + \delta
\sum_{v \in  \boldsymbol W ^c} d_v \| \gamma_{\des(v)} \|.
$$
The consistency condition is then obtained by studying when the first order expansion indeed has the correct sparsity pattern (for more precise statements and arguments, see \cite{fu}). We let denote $\gamma_{ \boldsymbol W }$ the solution of the previous problem, restricted to $\gamma_{{ \boldsymbol W }^c}=0$. We have:
$$  \gamma_{{ \boldsymbol W }}= \delta \boldsymbol \Sigma_{{ \boldsymbol W }{ \boldsymbol W }}^{-1} \Diag\left(   \textstyle
\sum_{v \in \anc(w)}   \frac{ d_v }{\| \boldsymbol\beta_{\des(v) \cap { \boldsymbol W }} \|} \right)_{w\in { \boldsymbol W }} \beta_{ \boldsymbol W }.
$$
Following the previous section, it is optimal if and only for all $\Delta \in { \boldsymbol W }^c$,
$$
\Delta_{{ \boldsymbol W }^c}^\top \boldsymbol\Sigma_{{ \boldsymbol W }^c { \boldsymbol W }} \gamma_{ \boldsymbol W } + 
 \delta
 \left(  \sum_{v \in { \boldsymbol W }^c} d_v \| \Delta_{\des(v)} \|\right) \geqslant 0.
$$
We let denote
$$A_{{ \boldsymbol W }^c}= \delta^{-1} \boldsymbol \Sigma_{{ \boldsymbol W }^c { \boldsymbol W }} \gamma_{ \boldsymbol W } = \boldsymbol\Sigma_{{ \boldsymbol W }^c { \boldsymbol W }}  \boldsymbol \Sigma_{{ \boldsymbol W }{ \boldsymbol W }}^{-1} \Diag\left(   \textstyle
\sum_{v \in \anc(w)}   \frac{ d_v }{\| \boldsymbol\beta_{\des(v) \cap { \boldsymbol W }} \|} \right)_{w\in { \boldsymbol W }} \boldsymbol\beta_{ \boldsymbol W }. $$
The condition for good pattern selection is that
for all $\Delta \in { \boldsymbol W }^c$,
$$
\Delta_{{ \boldsymbol W }_c}^\top A_{{ \boldsymbol W }^c}
+
  \sum_{v \in { \boldsymbol W }^c} d_v \| \Delta_{\des(v)} \|  \geqslant 0,
$$
which is exactly equivalent to $\| A_{{ \boldsymbol W }^c} \|^\ast \leqslant 1$, where $x \mapsto \| x \|^\ast$ is the dual norm of 
the norm $\Delta_{{ \boldsymbol W }^c} \mapsto \sum_{v \in { \boldsymbol W }^c} d_v \| \Delta_{\des(v)} \|$. This dual norm may be computed in closed form in the unstructured case, where $\des(v) = v$, and is equal to the $\ell^\infty$-norm. In general, it cannot be computed in closed form. However, we can give the following lower and upper bounds that lead to the desired propositions of the main paper.

We have:
$$ \sum_{v \in { \boldsymbol W }^c} d_v \| \Delta_{\des(v)} \| \leqslant 
\sum_{v \in { \boldsymbol W }^c}  \sum_{w \in \des(v) } d_v \| \Delta_{w} \|
= \sum_{w \in { \boldsymbol W }^c}  \left( \sum_{v \in \anc(v) \cap { \boldsymbol W }^c } d_v   \right)  \| \Delta_{w} \|,
$$
which leads to the upper bound $$\| x\|^\ast \leqslant \max_{w \in { \boldsymbol W }^c} \frac{ \| x_w\|}{\sum_{v \in \anc(v) \cap { \boldsymbol W }^c } d_v  }
$$

Moreover, we have:
\BEAS
\left( \sum_{v \in { \boldsymbol W }^c} d_v \| \Delta_{\des(v)} \| \right)^2
& = &  \sum_{v \in { \boldsymbol W }^c} \sum_{v' \in { \boldsymbol W }^c} d_v d_{v'} \| \Delta_{\des(v)} \| \| \Delta_{\des(v')} \| \\
& \geqslant & 
 \sum_{v \in { \boldsymbol W }^c} \sum_{v' \in { \boldsymbol W }^c} d_v d_{v'} \| \Delta_{\des(v)\cap \des(v') } \|^2 \\
& = & 
 \sum_{v \in { \boldsymbol W }^c} \sum_{v' \in { \boldsymbol W }^c}  \| \Delta_w\|^2 \sum_{w\in \des(v)\cap \des(v')}  d_v d_{v'}  \\ 
 & = & \sum_{w \in { \boldsymbol W }^c} \| \Delta_{w} \|^2 \sum_{v \in \anc(w) \cap { \boldsymbol W }^c} \sum_{v' \in \anc(w) \cap { \boldsymbol W }^c} d_v d_{v'} \\
 & = & \sum_{w \in { \boldsymbol W }^c} \| \Delta_{w} \|^2 \left( \sum_{v \in \anc(w) \cap { \boldsymbol W }^c}  d_{v} \right)^2.
 \EEAS
which leads to the lower bound:
 $$( \| x\|^\ast)^2 \geqslant \sum_{w \in { \boldsymbol W }^c} \frac{ \| x_w\|^2}{  \left( \sum_{v \in \anc(v) \cap { \boldsymbol W }^c } d_v  \right)^2 }.
$$

\label{sec:consistency-proof}

\bibliography{mikl}

\end{document}